\pdfoutput=1

\documentclass[11pt,table,dvipsnames]{article}

\usepackage[preprint]{acl}

\usepackage{times}
\usepackage{latexsym}
\usepackage{booktabs}
\usepackage[T1]{fontenc}

\usepackage[utf8]{inputenc}

\usepackage{microtype}

\usepackage{inconsolata}

\usepackage{graphicx}
\usepackage{colortbl}

\usepackage[most]{tcolorbox}

\usepackage{verbatim}
\usepackage{stfloats}
\usepackage{listings}

\newcounter{promptctr}
\renewcommand{\thepromptctr}{Prompt~\Alph{promptctr}}

\tcbset{
  bw:domain/.style={
    colback=white,
    colframe=black!40,
    coltitle=black,
    fonttitle=\bfseries\sffamily,   
    colbacktitle=examplebg,
    boxrule=0.5pt,
    titlerule=0pt,
    arc=2pt,
    left=4pt,right=4pt,top=4pt,bottom=4pt,
    toptitle=2pt,bottomtitle=2pt,
  }
}

\newcommand{\mytcbinputwide}[5]{
  \begin{figure*}[t]
  \centering
  \refstepcounter{promptctr}
  \phantomsection
  \begin{tcolorbox}[title={\thepromptctr: #2},#4,width=\textwidth,enhanced]
    \lstinputlisting{#1}
    \label{#5}
  \end{tcolorbox}
  \vspace{-4pt}
  \end{figure*}
}

\newcommand{\promptref}[1]{%
  \hyperref[#1]{\ref*{#1}}%
}

\newcommand{\promptrefp}[1]{%
  \hyperref[#1]{\ref*{#1} (p.~\pageref*{#1})}%
}

\title{Rethinking Information Synthesis in Multimodal Question Answering \\ A Multi-Agent Perspective}

\usepackage{xspace}
\newcommand{\methodName}{{\sc MAMMQA~}\xspace}
\usepackage{enumitem}

\usepackage{mdframed}

\definecolor{examplebg}{RGB}{220,245,245}

\newtcolorbox{exampleblock}{
  enhanced,
  colback=examplebg,
  colframe=examplebg,   
  boxrule=0pt,
  arc=2mm,
  outer arc=2mm,
  left=8pt,
  right=8pt,
  top=6pt,
  bottom=6pt,
  boxsep=0pt,
  before skip=8pt,
  after skip=8pt,
  sharp corners=all,
  breakable
}

\author{
Tejas Anvekar\thanks{contributed equally} \quad 
Krishna Singh Rajput\footnotemark[1] \quad 
Chitta Baral \quad 
Vivek Gupta\thanks{primary supervisor} \\
Arizona State University \\
\texttt{\{tanvekar,krajput5,chitta,vgupt140\}@asu.edu} \\
\href{https://coral-lab-asu.github.io/MAMMQA/}{https://coral-lab-asu.github.io/MAMMQA/}
}

\begin{document}
\maketitle

\begin{abstract}
Recent advances in multimodal question answering have primarily focused on combining heterogeneous modalities or fine-tuning multimodal large language models. While these approaches have shown strong performance, they often rely on a single, generalized reasoning strategy, overlooking the unique characteristics of each modality ultimately limiting both accuracy and interpretability. To address these limitations, we propose \methodName, a multi-agent QA framework for multimodal inputs spanning text, tables, and images. Our system includes two Visual Language Model (VLM) agents and one text-based Large Language Model (LLM) agent. The first VLM decomposes the user query into sub-questions and sequentially retrieves partial answers from each modality. The second VLM synthesizes and refines these results through cross-modal reasoning. Finally, the LLM integrates the insights into a cohesive answer. This modular design enhances interpretability by making the reasoning process transparent and allows each agent to operate within its domain of expertise. Experiments on diverse multimodal QA benchmarks demonstrate that our cooperative, multi-agent framework consistently outperforms existing baselines in both accuracy and robustness.
\end{abstract}

\section{Introduction}
Multimodal question answering (MMQA) aims to answer complex queries by jointly reasoning over text, tables, and images, reflecting real‐world information needs in domains such as scientific analysis, business intelligence, and education \citep{talmor2021multimodalqa,Hannan_Jain_Bansal_2020}. Early MMQA systems typically linearized tables or generated image captions to cast all inputs into a text‐only format, feeding them into pretrained text‐only models \citep{luo2023unifying,chen-etal-2020-hybridqa,chen2021ottqa}. While effective under certain settings, these unified approaches often obscure the unique structure and semantics of each modality, leading to degradation when inputs are missing or when fine‐grained visual and tabular cues are critical.

Recent advances in prompt‐based reasoning have unlocked zero‐shot and few‐shot capabilities in large language models. Chain‐of‐Thought (CoT) prompting \citep{wei2022chain} and its multimodal extensions \citep{zhang2023multimodal,zheng2023ddcot} guide a single LLM to generate intermediate steps, improving factual accuracy. Tree‐of‐Thoughts (ToT) further introduces search over multiple reasoning branches \citep{10.5555/3666122.3666639}. However, these monolithic strategies still treat the LLM as a black box, entangling modality‐specific extraction with cross‐modal synthesis, which can obscure errors, hinder interpretability, and induce hallucinations when faced with ambiguous or partial inputs.

\begin{figure}[t]
    \centering
    \includegraphics[width=1.0\linewidth]{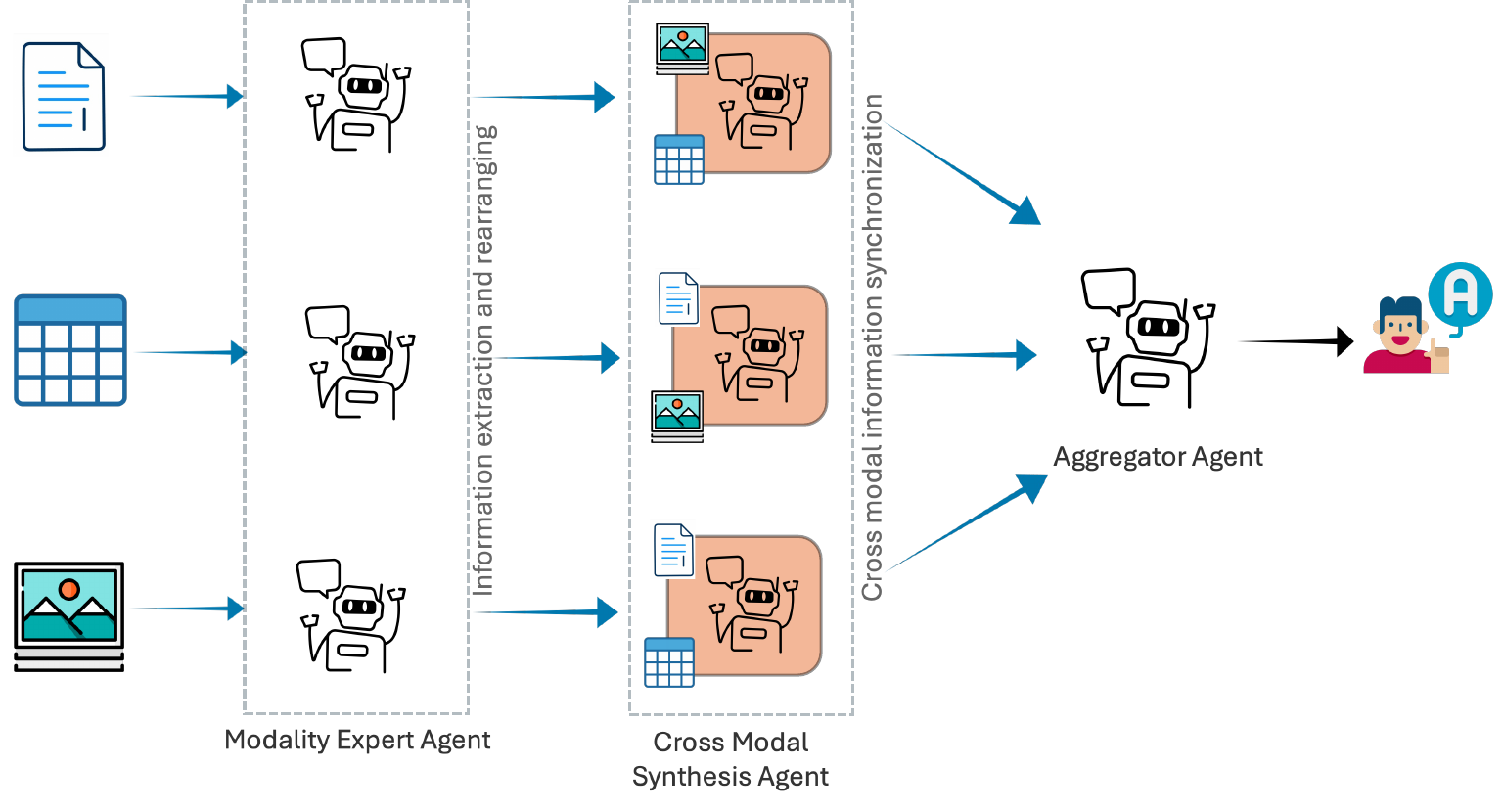}
    \vspace{-1.0em}
    \caption{Depicting Illustration for our proposed \methodName, with three agents: 1) Modality Expert, that extracts modality specific insights; 2) Cross Modal Systhesis Agent, that synchronises information across modalities with insights from Modality Expert; 3) Aggregator Agent, that ground the answer using extracted cross modal information.}
    \vspace{-1.5em}
    \label{fig:architecture}
\end{figure}

By contrast, multi‐agent and ensemble techniques in NLP have demonstrated that specialized experts can collaborate to improve both accuracy and robustness \citep{chen2023reconcile,puerto-etal-2023-metaqa}. Yet, such architectures remain underexplored in the multimodal setting. We identify a key opportunity: decoupling modality‐specific evidence extraction from cross‐modal integration and final answer adjudication can both leverage domain‐specific strengths and provide transparent, verifiable reasoning traces.

In this work, we introduce \methodName, a fully prompt‐driven, multi‐agent framework for MMQA that dynamically allocates three types of agents modality experts, cross‐modal synthesizers, and a consensus aggregator to decompose and solve complex queries without any fine‐tuning. Our core contributions are:
\begin{itemize}
  \item \textbf{\methodName}: A fully prompt-driven, multi-agent MMQA framework that splits reasoning into three interpretable stages modality experts, cross-modal synthesis, and evidence-grounded aggregation without any fine-tuning.
  \item \textbf{Unified, role-consistent agents}: A single prompt template reused across text, table, and image experts, enabling dynamic activation, efficient inference, and transparent error tracing.
  \item \textbf{State-of-the-art zero-shot performance}: Outperforms CoT, CapCoT, and ToT baselines and matches or exceeds several fine-tuned models on \textsc{MultiModalQA} and \textsc{ManyModalQA}, across both proprietary and open-source LLMs.
  \item \textbf{Robustness and calibration}: Static agents beat dynamic search methods (e.g.\ ToT) by over 10 \%, maintain faithfulness under noise and irrelevant context, and avoid hallucinations via evidence-based abstention.
\end{itemize}

By structuring MMQA as a pipeline of specialized agents, \methodName not only achieves high accuracy but also provides end‐to‐end transparency and graceful failure modes in the face of ambiguous or incomplete inputs key properties for real‐world deployment.

\section{Our \methodName Framework}
\label{sec:method}

We propose \textbf{\methodName}, a Multi-Agent Multi-modal Question Answering framework designed to address core challenges in MMQA, such as modality ambiguity, fragmented evidence, and hallucinated reasoning. Rather than relying on monolithic prompting or fine-tuned end-to-end models, \methodName adopts a structured, agent-based architecture that decomposes the reasoning process into interpretable, well-defined steps. Each agent specializes in a narrow subtask, enabling systematic insight extraction, targeted cross-modal synthesis, and final answer adjudication through consensus.

\subsection{Motivation and Design Principles}

Multimodal question answering requires models to accurately interpret and integrate information across diverse modalities. However, conventional prompting approaches often struggle with modality disambiguation and fail to coordinate evidence coherently. To address this, our framework is inspired by the structure of expert committees, where domain specialists independently contribute insights that are later synthesized into a final decision. \methodName is guided by three core principles. Each agent is assigned a focused, well-defined role either modality-specific analysis, cross-modal synthesis, or answer aggregation mirroring real-world task delegation. The reasoning process is decomposed into a multi-stage pipeline that progresses from factual extraction to cross-modal interpretation and ultimately to consensus. The entire framework is prompt-driven, leveraging pre-trained language models without requiring any task-specific fine-tuning.

\subsection{Overview of the \methodName Architecture}

The \methodName framework consists of three sequential stages: modality-specific insight extraction, cross-modality synthesis and reasoning, and final answer aggregation. Each stage employs a set of pre-trained language models acting as agents, each prompted with a structured task definition. The number of active agents in the system dynamically adjusts depending on the available input modalities, ranging from five in the bi-modal case to seven for full tri-modal inputs. All agents interact through textual interfaces, ensuring that the pipeline remains fully modular and interpretable, best illustrated by~\autoref{fig:architecture}.

\subsection{Stage I: Modality Expert Agent}
The first stage applies a unified modality expert agent to each available input modality text, table, and image. Although executed independently for each modality, the same underlying system prompt is used across all instances; the only variation lies in the input modality. This agent is not tasked with answering the question but instead focuses on extracting key information relevant to the query, such as factual details, temporal markers, and contextual cues. It also flags gaps or ambiguities in the input that may impede complete reasoning. The agent's output is structured using a consistent templating format, producing modality-specific insights that form the foundation for the next reasoning stage.

\subsection{Stage II: Cross-Modality Synthesis Agent}

In the second stage, the same cross-modality synthesis agent is invoked once for each modality acting as the anchor context. Each instance of the agent takes as input the insights from a single modality-specific agent in Stage I, the raw data from the remaining two modalities, and the original question. This setup allows the agent to synthesize information across modalities while maintaining a consistent perspective grounded in one primary modality. The agent generates a complete answer to the question, supports it with structured reasoning, and identifies any remaining uncertainties. As with the first stage, the output adheres to a templated format comprising extracted insights, intermediate sub-answers, and a final answer. These outputs are then passed to the final aggregation stage.

\subsection{Stage III: Aggregator Agent}

The final stage introduces an aggregation agent responsible for synthesizing the outputs of the three synthesis agents. Crucially, this agent does not have access to the raw inputs. Instead, it receives only the generated responses and reasoning from the prior stage, along with the original question. Its task is to resolve disagreements, consolidate consistent answers, and produce a final, justified response. The agent follows a hierarchical decision process: it first checks for answer consistency among the agents; if two or more agree and present clear reasoning, it adopts that answer. If two agents express uncertainty while the third provides a confident, well-supported answer, the agent selects the confident one. In cases where all three answers differ, it evaluates them based on the clarity, coherence, and strength of their rationale, ultimately choosing the most convincing explanation. The final output includes both the chosen answer and a transparent summary of the reasoning process that led to its selection.

\subsection{Advantages of the Multi-Agent Prompting Framework}

The agent-based design of \methodName delivers interpretable and robust multimodal question answering. Its transparency stems from traceable intermediate steps, and robustness is achieved through redundant synthesis agents for answer cross-verification. Crucially, \methodName scales across LLM sizes and domains using only prompting, avoiding fine-tuning. This modularity supports generalization and faithful data grounding, establishing a framework for enhanced reliability, accuracy, and interpretability within \textsc{MultiModalQA}.

\section{Experiments}
\label{sec:Exp}

\begin{table*}[t]
\centering
\begin{tabular}{lccccccc}
\hline \hline
\textbf{Modality} & \textbf{Img} & \textbf{Tb | Img} & \textbf{Tb | Txt} & \textbf{Tb} & \textbf{Txt | Img} & \textbf{Txt} & \textbf{Total} \\ \hline
\multicolumn{8}{c}{\cellcolor[HTML]{EFEFEF}\textit{OpenAI 4o Mini}}                                                           \\
CoT& 33.15          & 53.81          & 66.67          & \underline{84.55}          & 55.95          & \underline{77.67}          & 64.60          \\
CapCoT& 53.91          & \underline{64.98}          & \underline{69.05}          & 84.14          & \textbf{61.90} & 77.33          & \underline{70.39}          \\
ToT& \underline{54.97}          & 63.35          & 64.37          & 67.70          & \underline{61.11}          & 69.65          & 64.88          \\
\textbf{Ours}   & \textbf{61.31} & \textbf{70.30} & \textbf{81.58} & \textbf{89.16} & 59.75          & \textbf{85.57} & \textbf{76.37} \\
\multicolumn{8}{c}{\cellcolor[HTML]{DAE8FC}\textit{Gemini 1.5-Flash 8B}}                                                      \\
CoT& 47.41          & \underline{53.38}          & \textbf{58.88} & 74.73          & \textbf{46.43} & \underline{72.82}          & \underline{62.16}          \\
CapCoT& \underline{47.84}          & 50.02          & 55.87          & \underline{74.88}          & 39.29          & 72.42          & 60.66          \\
ToT& 36.93          & 43.06          & 52.32          & 53.72          & 33.33          & 70.61          & 53.10          \\
\textbf{Ours}   & \textbf{51.23} & \textbf{54.12} & \underline{57.42}          & \textbf{83.69} & \underline{42.86}          & \textbf{79.47} & \textbf{65.84} \\
\multicolumn{8}{c}{\cellcolor[HTML]{D6FFD6}\textit{\texttt{Qwen} 2.5 VL 7B Instruct}}                                                  \\
CoT& 29.11          & 32.58          & 30.66          & 38.75          & 17.86          & 38.28          & 33.84          \\
CapCoT& 48.10          & \underline{53.94}          & \underline{60.56}          & \underline{71.52}          & \underline{41.67}          & \underline{71.31}          & \underline{61.54}          \\
ToT& \textbf{55.90} & 47.82          & 52.50          & 60.83          & 41.64          & 64.44          & 57.12          \\
\textbf{Ours}   & \underline{50.74}          & \textbf{55.88} & \textbf{63.68} & \textbf{81.35} & \textbf{53.26} & \textbf{80.51} & \textbf{67.56} \\
\multicolumn{8}{c}{\cellcolor[HTML]{FFFDC1}\textit{\texttt{Qwen} 2.5 VL 3B Instruct}}                                                  \\
CoT& 11.86          & 23.71          & 22.14          & 32.25          & 14.29          & 25.52          & 23.15          \\
CapCoT& \textbf{48.10} & 42.08          & \underline{47.08}          & \textbf{64.94} & \textbf{39.29} & 65.04          & \textbf{53.98} \\
ToT& \underline{42.01}          & \textbf{43.65} & \textbf{48.40} & 52.57          & 33.74          & \underline{66.51}          & \underline{52.91}          \\
\textbf{Ours}   & 33.73          & \underline{43.10}          & 45.33          & \underline{62.29}          & \underline{35.52}          & \textbf{67.73} & 52.12          \\ \hline \hline
\end{tabular}%
\caption{Quantitative Analysis on \textsc{MultiModalQA}  dataset}
\label{tab:multimodal}
\end{table*}

\begin{table}[t]
\small
\centering
\begin{tabular}{lcccc}
\hline \hline
\textbf{Methods}                      & \textbf{Text}  & \textbf{Table} & \textbf{Image} & \textbf{Total} \\ \hline
Human                        & \textit{92.00}             & \textit{89.60}           & \textit{94.00}             & \textit{91.60}           \\
Voting                       & 23.70           & 22.90           & 15.50          & 21.10           \\
MMQA                  & 48.60           & 40.40           & 27.20           & 39.70           \\
MMQA\textbf{\textsuperscript{\dag}} & 59.30           & 46.30           & 29.00             & 46.30           \\
\multicolumn{5}{c}{\cellcolor[HTML]{C3FFFD}\textit{UniMMQA Finetuned T5 Model}}                          \\
Base               & 46.60           & 60.70           & 30.20           & 45.40           \\
Large              & 48.50           & 67.50           & 34.90           & 50.00             \\
3B                         & 49.80           & 58.30           & 40.90           & 52.10           \\
\multicolumn{5}{c}{\cellcolor[HTML]{EFEFEF}\textit{OpenAI 4o-mini}}                              \\
CoT                          & 87.20           & \underline{94.23}          & 57.33          & 81.21          \\
CoT\textbf{\textsuperscript{*}} & \cellcolor[HTML]{FFBABA}68.22 & \cellcolor[HTML]{FFBABA}70.51 & \cellcolor[HTML]{FFBABA}59.42 & \cellcolor[HTML]{FFBABA}66.54 \\
CapCoT                       & \underline{87.68}          & 94.05          & 68.26          & 84.41          \\
ToT                          & 84.94          & 93.19          & \underline{72.90}          & \underline{84.70}          \\
\textbf{Ours}                         & \textbf{92.50} & \textbf{96.78} & \textbf{78.02} & \textbf{89.90} \\
\multicolumn{5}{c}{\cellcolor[HTML]{DAE8FC}\textit{Gemini 1.5-Flash 8B}}                         \\
CoT                          & 86.05          & \underline{91.52}          & \underline{68.77}          & \underline{82.81}          \\
CoT\textbf{\textsuperscript{*}} & \cellcolor[HTML]{FFBABA}54.93 & \cellcolor[HTML]{FFBABA}61.15 & \cellcolor[HTML]{FFBABA}34.77 & \cellcolor[HTML]{FFBABA}51.41 \\
CapCoT                       & 85.74          & 91.40          & 63.14          & 81.34          \\
ToT                          & \underline{86.08}          & 86.81          & 62.81          & 79.80          \\
\textbf{Ours}                         & \textbf{89.76} & \textbf{94.52} & \textbf{77.33} & \textbf{87.91} \\
\multicolumn{5}{c}{\cellcolor[HTML]{D6FFD6}\textit{\texttt{Qwen} 2.5 VL 7B Instruct}}                     \\
CoT                          & 59.84          & 68.71          & 45.47          & 58.87          \\
CoT\textbf{\textsuperscript{*}} & \cellcolor[HTML]{FFBABA}61.80 & \cellcolor[HTML]{FFBABA}66.73 & \cellcolor[HTML]{FFBABA}54.53 & \cellcolor[HTML]{FFBABA}61.46 \\
CapCoT                       & \underline{83.50}          & \underline{92.86}          & \underline{71.07}          & \underline{83.41}          \\
ToT                          & 81.95          & 90.41          & 69.29          & 81.89          \\
\textbf{Ours}                         & \textbf{87.11} & \textbf{96.31} & \textbf{77.56} & \textbf{87.61} \\
\multicolumn{5}{c}{\cellcolor[HTML]{FFFDC1}\textit{\texttt{Qwen} 2.5 VL 3B Instruct}}                     \\
CoT                          & 70.08          & 75.61          & 50.70          & 66.54          \\
CoT\textbf{\textsuperscript{*}} & \cellcolor[HTML]{FFBABA}58.77 & \cellcolor[HTML]{FFBABA}64.55 & \cellcolor[HTML]{FFBABA}59.51 & \cellcolor[HTML]{FFBABA}58.77 \\
CapCoT                       & 80.79          & \underline{91.38}          & 67.13          & \underline{80.63}          \\
ToT                          & \underline{82.66}          & 86.14          & \underline{68.11}          & 80.42          \\
\textbf{Ours}                         & \textbf{88.79} & \textbf{94.90} & \textbf{72.67} & \textbf{86.37} \\ \hline \hline
\end{tabular}%


\caption{Quantitative results on the \textsc{ManyModalQA} dataset. 
Superscript $\dagger$ denotes the oracle setting, while * indicates the no-context (open-book QA) variant. Red highlights mark cases where CoT fails to abstain from answering without context, and occasionally outperforms its baseline in the [No Image] setting, suggesting potential data leakage.}

\label{tab:manymodal}
\vspace{-1.0em}
\end{table}

We evaluate the effectiveness of our method on the Multimodal Question Answering (MMQA) task using exact match, demonstrating superior performance compared to prior state-of-the-art approaches, including UniMMQA~\cite{luo2023unifying}, AutoRouting~\cite{talmor2021multimodalqa}, ImplicitDecomp~\cite{talmor2021multimodalqa}, Binder~\cite{unknown}, SKURG~\cite{inproceedings}, PERQA~\cite{yang2023progressiveevidencerefinementopendomain}, Solar~\cite{yu2023unifiedlanguagerepresentationquestion}, UniRaG~\cite{sharifymoghaddam-etal-2025-unirag}, AETGA~\cite{10.1145/3664647.3681479}, and PReasM-Large~\cite{DBLP:journals/corr/abs-2107-07261}. Additionally, we benchmark against standard in-context prompting baselines such as Chain-of-Thoughts~\cite{wei2022chain} (CoT), Image-Captioning + CoT (CapCoT), and Tree-of-Thoughts~\cite{10.5555/3666122.3666639} (ToT), across both proprietary (\texttt{gpt-4o-mini}~\cite{openai_gpt4omini_2024}, \texttt{Gemini-1.5-flash-8B}~\cite{geminiteam2024gemini15unlockingmultimodal}) and open-source models (\texttt{Qwen2.5}-\texttt{VL}-\texttt{Instruct}-\texttt{7B/3B}~\cite{qwen2025qwen25technicalreport}). We further conduct endurance tests to assess the robustness of our approach under challenging scenarios. \textbf{Note:} All experiments involving \texttt{Qwen} models were conducted locally on a system equipped with 8× NVIDIA H200 GPUs. 

\paragraph{Datasets} We evaluate our approach on two prominent benchmark datasets designed to test key reasoning capabilities in multimodal question answering (MMQA).

\textsc{ManyModalQA}~\cite{Hannan_Jain_Bansal_2020} contains 10,190 questions involving text, images, and tables distributed across 2,873 images, 3,789 passages, and 3,528 tables. The dataset is intentionally constructed with ambiguous questions where the relevant modality is not explicitly indicated. This design tests a model's ability to perform modality disambiguation and selectively retrieve relevant information. It highlights whether the model can reason about what modality is needed and how to integrate it effectively. With 2,036 training and 3,055 development examples, it serves as a strong benchmark for evaluating modality selection under uncertainty.

\textsc{MultiModalQA}~\cite{talmor2021multimodalqa} consists of 29,918 question-answer pairs across multiple modalities, with a significant 35.7\% of questions requiring cross-modal reasoning that is, combining evidence from different types of inputs. This dataset assesses a model’s ability to integrate heterogeneous information and reason jointly across modalities, not just within a single source. It is divided into training (23,817), development (2,442), and test (3,660) splits, and is particularly suited for evaluating compositional reasoning and information fusion in complex multimodal contexts. 

\paragraph{Baselines}
To comprehensively evaluate our method, we compare it against several strong baselines spanning both finetuned and prompting-based approaches for multimodal question answering. 

\textit{Finetuned Baseline.} 
\textbf{UniMMQA}~\cite{luo-etal-2023-unifying} serves as a T5~\cite{2020t5}-based finetuned baseline and represents a strong state-of-the-art model for MMQA. It is trained with supervised signals across modalities, enabling robust cross-modal representation learning.

\textit{Prompting-Based Agents.} 
Our proposed prompting strategies adopt an agent in-context learning setup. Each agent specializes in a modality and collaborates to perform modality disambiguation and reasoning, allowing the model to retrieve, decompose, and integrate evidence dynamically without any parameter updates.

\textit{Reasoning Variants.} 
\textbf{Chain-of-Thought (CoT)} guides the model to generate intermediate reasoning steps from the raw question and context.

\textbf{CapCoT} enhances CoT by incorporating detailed image and table captions generated via Gemini-2.0-Flash, providing richer modality cues in textual form.

\textbf{Tree-of-Thought (ToT)} extends CapCoT by simulating multiple reasoning paths using a depth-first search (DFS) strategy over caption-augmented inputs, introducing structured exploration for better answer synthesis.

\begin{table}[t]
\small
\centering
\begin{tabular}{lccc}
\hline \hline
\textbf{Model} & \textbf{Single} & \textbf{Multi} & \textbf{Overall} \\ \hline
\multicolumn{4}{c}{\cellcolor[HTML]{EFEFEF}\textit{\textbf{Finetuned Models}}} \\
AutoRouting            & 51.7          & 34.2          & 44.7          \\
ImplicitDecomp         & 51.6          & 44.6          & 48.8          \\
Binder                 & -            & -            & 51.0          \\
SKURG                  & 66.1          & 52.5          & 59.8          \\
PERQA                  & 69.7          & 54.7          & 62.8          \\
Solar                  & 69.7          & 55.5          & 59.8          \\
UniRaG                 & \textbf{71.7}          & \underline{62.3}          & 67.4          \\
AETGA                  & \underline{69.8}          & \textbf{64.7}          & 68.8          \\
PReasM L           & -            & -            & 59.0          \\
MMQA-T5 L         & -            & -            & 57.9          \\
UniMMQA (T5 B)      & -            & -            & 67.9          \\
UniMMQA (T5 L)     & -            & -            & \underline{71.3}          \\
UniMMQA (T5 3B)        & -            & -            & \textbf{75.5}          \\

\multicolumn{4}{c}{\cellcolor[HTML]{D6FFD6}\textit{\textbf{Zero-Shot Models}}} \\
CoT \texttt{Qwen} 3B            & 23.75         & 22.24         & 23.15         \\
CoT \texttt{Qwen} 7B            & 36.07         & 30.91         & 33.84         \\
Our Agent 3B           & \underline{57.72}         & \underline{43.39}         & \underline{52.12}         \\
\textbf{Our Agent 7B}  & \textbf{73.16} & \textbf{58.93} & \textbf{67.56} \\ \hline \hline
\end{tabular}%
\caption{Comparison of models on \textsc{MultiModalQA}  dataset across single-modality, multi-modality, and overall performance. Note: B depicts Base Model and L depicts Large model.}
\label{tab:multibaseline_overall}
\vspace{-1.0em}
\end{table}

\paragraph{LLM Configurations.} All models, both proprietary and open-source, are evaluated with a \texttt{temperature} of 0.3 and \texttt{top-p} of 0.7 to ensure consistent generation behavior. We employ the \texttt{framework}\footnote{\url{https://github.com/openai/openai-agents-python}} to implement Tree-of-Thought (ToT) agents. Our proposed agents operate in a static and synchronous manner, removing the need for asynchronous execution. The corresponding prompts are detailed in Appendix~\ref{sec:appendix_prompts}: \promptref{prompt:AS1} presents the Modality Expert Agent prompt, \promptref{prompt:AS2} illustrates the Cross-Modality Agent prompt, and \promptref{prompt:AS3} provides the Aggregator prompt.

\subsection{Comparison with State-of-the-Art}

\paragraph{A. \textsc{MultiModalQA} Results.} As shown in ~\autoref{tab:multibaseline_overall}, our method consistently outperforms prompting-based baselines across both proprietary and open-source models. On \texttt{Qwen2.5-VL-7B}, our agentic method achieves \textbf{76.37\%}, surpassing CapCoT (\textbf{+5.98\%}) and ToT (\textbf{+11.49\%}). Largest gains are observed in cross-modal settings like \textit{[table, text]} (+17.21\%) and \textit{image} (+7.4\%), underscoring the benefit of grounded signal integration. On Gemini-1.5 Flash 8B, we achieve \textbf{65.84\%}, outperforming CoT by \textbf{+3.68\%}. While prompting methods show less spread here, our model remains notably better in \textit{text} (+6.65\%) and \textit{table} (+8.81\%), showing advantages in structurally complex inputs. 

\paragraph{Open-Source Scaling (\texttt{Qwen}).} Our method yields strong results even on small open-source models. On \texttt{Qwen2.5-VL-3B}, it achieves \textbf{52.12\%}, beating CoT by \textbf{+28.97\%} and closely trailing ToT (\textbf{-0.79\%}) despite significantly lower computation. On \texttt{Qwen2.5-VL-7B}, our method reaches \textbf{67.56\%}, outperforming CapCoT (\textbf{+6.02\%}), ToT (\textbf{+10.44\%}), and CoT (\textbf{+33.72\%}). These gains are especially prominent in multi-hop, hybrid modality tasks e.g., \textit{[text, image]} (+90.53\%) and \textit{[table, text]} (+40.48\%). 

\paragraph{Model Scaling.} Increasing model size from 3B to 7B brings a substantial \textbf{+29.62\%} gain in overall performance. Improvements are concentrated in \textit{image} (+50.43\%) and \textit{[text, image]} (+49.95\%) modalities, suggesting that increased capacity amplifies our agentic system’s ability to perform complex, cross-modal reasoning.

\paragraph{Efficiency Over Larger Baselines.} Despite its smaller size, our 3B model surpasses \texttt{Qwen-7B} CoT on multiple modalities \textit{text} (+76.94\%), \textit{[table, image]} (+32.31\%) with an overall gain of \textbf{+54.02\%}. This highlights the architectural efficiency of structured agentic reasoning over naive pattern-based prompting.

\paragraph{B. \textsc{ManyModalQA} Results.} Our method generalizes well on this more challenging benchmark. On \texttt{Qwen2.5-VL-7B}, we score \textbf{89.90\%}, outperforming CoT by \textbf{+8.69\%}, ToT by \textbf{+5.20\%}, and CapCoT by \textbf{+5.49\%}. The largest gains occur in visual reasoning tasks e.g., \textit{image} (+5.12\%). On Gemini 1.5-Flash 8B, we obtain \textbf{87.91\%}, with consistent improvements across modalities: \textbf{+5.10\%} over CoT, \textbf{+8.11\%} over ToT, and \textbf{+6.57\%} over CapCoT. Our method shows stronger synergy between vision and language compared to captioning-heavy baselines. 

\paragraph{Open-Source Performance On \texttt{Qwen2.5-VL-7B}} we achieve \textbf{87.61\%}, outperforming ToT (\textbf{+5.72\%}) and CapCoT (\textbf{+4.20\%}). Notably, we also surpass Gemini 1.5-8B in overall score (+0.30\%) and structured modalities like \textit{text} (+4.32\%) and \textit{table} (+5.38\%). On \texttt{Qwen2.5-VL-3B}, our model reaches \textbf{86.37\%}, significantly ahead of CoT (\textbf{+19.83\%}) and CapCoT (\textbf{+5.74\%}), and even outperforming Gemini-8B on both total score (+1.56\%) and image (+8.56\%).

\paragraph{Model Scaling.} Scaling from \texttt{Qwen} 3B to 7B gives a modest \textbf{+1.24\%} overall, but a pronounced boost in visual reasoning (\textbf{+6.72\%}). This suggests that while our architecture is already strong at 3B, larger models especially enhance performance on ambiguous, visually grounded questions.

\paragraph{Finetuned vs. Zero-shot.} Despite being zero-shot, our 7B method outperforms several finetuned baselines e.g., SKURG (59.8\%), Solar (59.8\%) and rivals AETGA (68.8\%) and UniRaG (67.4\%). Compared to \texttt{Qwen-7B} CoT, we observe large gains in both single-modality (\textbf{+37.09\%}) and multi-modality (\textbf{+28.02\%}) settings. Even our 3B variant exceeds \texttt{Qwen-7B} CoT by \textbf{+18.2\%}, reaffirming that architecture not just size drives robust performance.

\subsection{Robustness Analysis}

\paragraph{\methodName Mislabeling Robustness.} In our experiments with the \textsc{MultiModalQA}  dataset, we evaluated multiple LLMs such as GPT and \texttt{Qwen} using baseline methods CoT, CapCoT, Tree-of-Thoughts, and our agentic method. During the analysis, we discovered discrepancies in the ground truth labels of the dataset, such as typos and outdated factual information. For instance, an answer labeled movie name ``\textit{laughin}" should have been ``\textit{laughing}." This led to certain models, particularly CoT, memorizing and reproducing these incorrect labels, thereby inflating their performance metrics artificially.

\begin{table}[!ht]
\small
\centering
\begin{tabular}{lcc}
\hline \hline
\textbf{Model (\texttt{Qwen} 7B)} & \textbf{Old} & \textbf{New} \\ \hline
TreeOfThoughts      & 57.12        & 59.06 {\color{ForestGreen}(+1.94)}\\
CoT                 & 33.84        & 35.05 {\color{ForestGreen}(+1.21)}               \\
CapCoT           & \underline{61.54}        & \underline{64.52} {\color{ForestGreen}(+2.98)}               \\
OurAgent            & \textbf{67.56} & \textbf{71.58} {\color{ForestGreen}(+\textbf{4.02})}      \\ \hline
\textbf{Model (\texttt{Qwen} 3B)} & \textbf{Old} & \textbf{New} \\ \hline
TreeOfThoughts      & 52.91        & 54.36 {\color{ForestGreen}(+1.45)}               \\
CoT                 & 23.15        & 24.07 {\color{ForestGreen}(+0.92)}               \\
CapCoT           & \underline{53.98}        & \underline{56.16} {\color{ForestGreen}(+2.18)}               \\
OurAgent            & \textbf{52.12} & \textbf{55.24} {\color{ForestGreen}(+\textbf{3.12})}      \\ \hline \hline
\end{tabular}%
\caption{Performance improvements with lable correction across model sizes on \textsc{MultiModalQA} .}
\label{tab:gain_projection}
\end{table}

After correcting these labels, we observed that the performance of our agentic method improved significantly more than the baseline methods as depicted in ~\autoref{tab:gain_projection}. This highlights the \textit{robustness of our approach in extracting and synthesizing information from multiple modalities and grounding it accurately, even when faced with noisy or inconsistent data}. This correction process ultimately underscores the efficacy of our model in real-world scenarios.

\paragraph{\methodName Pertubations Robustness.} ~\autoref{tab:robustness_comparison} evaluates model robustness under two text-level perturbations: (1) sentence or paragraph shuffling and (2) injection of irrelevant context. In the \textit{Text Shuffle} setting, baseline methods like TreeOfThoughts (7B: \textcolor{red}{-42.21\%}, 3B: \textcolor{red}{-6.97\%}) and CapCoT (7B: \textcolor{red}{-39.11\%}, 3B: \textcolor{red}{-8.82\%}) exhibit substantial drops but still attempt to answer suggesting reliance on memorized question patterns. In contrast, \methodName exhibits steep performance drops (7B: \textbf{-91.24\%}, 3B: \textbf{-85.30\%}), suggesting that it fails gracefully under broken contextual grounding, thereby reducing the risk of hallucinated answers.

\begin{table}[!ht]
\small
\centering
\resizebox{0.95\linewidth}{!}{%
\begin{tabular}{lccc}
\hline \hline
\textbf{Model (7B)} & \textbf{Original} & \textbf{Text Shuffle} & \textbf{Irrelevant Context}  \\ \hline
TreeOfThoughts & 57.12 & 33.01 ({\color{red}-42.21\%}) & 52.45 ({\color{red}-08.18\%}) \\
CoT            & 33.84 & 31.18 ({\color{red}-07.86\%})  & 29.54 ({\color{red}-12.71\%})  \\
CapCoT      & \underline{61.54} & \underline{37.47} ({\color{red}-39.11\%}) & \underline{55.39} ({\color{red}-09.99\%})   \\
OurAgent       & \textbf{67.56} & \textbf{05.92} ({\color{red}-91.24\%}) & \textbf{63.74} ({\color{red}-05.65\%})  \\ \hline
\textbf{Model (3B)} & \textbf{Original} & \textbf{Text Shuffle} & \textbf{Irrelevant Context}  \\ \hline
TreeOfThoughts & 52.91 & \underline{49.22} ({\color{red}-06.97\%})  & 47.11 ({\color{red}-10.96\%})  \\
CoT            & 23.15 & 20.48 ({\color{red}-11.53\%}) & 19.62 ({\color{red}-15.25\%})  \\
CapCoT      & \underline{53.98} & \underline{49.22} ({\color{red}-08.82\%})  & \underline{47.12} ({\color{red}-12.71\%})  \\
OurAgent       & \textbf{52.12} & \textbf{07.66} ({\color{red}-85.30\%}) & \textbf{48.05} ({\color{red}-07.81\%}) \\
\hline \hline
\end{tabular}%
}
\caption{Robustness of different reasoning strategies under perturbations across model sizes.}
\label{tab:robustness_comparison}
\end{table}

In the \textit{Irrelevant Context} setting, where unrelated text is appended, OurAgent remains the most stable (7B: \textbf{-5.65\%}, 3B: \textbf{-7.81\%}) compared to TreeOfThoughts (7B: \textcolor{red}{-8.18\%}, 3B: \textcolor{red}{-10.96\%}) and CapCoT (7B: \textcolor{red}{-9.99\%}, 3B: \textcolor{red}{-12.71\%}). This demonstrates that while OurAgent avoids over-committing in incoherent contexts, it retains robustness when faced with extraneous information underscoring its grounding-driven reasoning approach.
\vspace{-2.0em}

\paragraph{\methodName Calibration Robustness.} Chain-of-Thought (CoT) prompting, while effective in unimodal text reasoning, fails to generalize reliably in multimodal contexts. On \textsc{MultiModalQA}, CoT consistently produces high-confidence yet unfaithful answers when modality-specific evidence is absent intentionally. This prompts an important question: \textit{Can LLMs, when operating under the \methodName{} framework, refrain from answering when provided with incomplete inputs?} As depicted in ~\autoref{tab:manymodal}, on \texttt{Qwen-7B}, CoT achieves 58.87\%, but this rises to 61.46\% in a no-context setting (CoT$^*$) indicating a reliance on pretraining priors rather than grounded inference.

This behavior suggests that CoT ``recalls" plausible reasoning paths learned during training rather than ``inferring" from the input analogous to a language model predicting the next sentence in a familiar story, even when the plot doesn’t match. In contrast, our agent-based architecture enforces structured, evidence-grounded reasoning. The Modality Expert Agent first extracts information independently from text, table, and image inputs. A Cross-Modality Expert then integrates these signals with consistency checks. Crucially, if no relevant evidence is found, these agents abstain from answering propagating that abstention to the Aggregator, which itself is blind to the original question. This ensures that the final output is generated only when sufficient grounded evidence exists.

This setup explicitly separates extraction from generation, reducing hallucinations and enforcing cross-modal faithfulness. As a result, our method achieves 89.90\% on OpenAI \texttt{Qwen2.5-VL-7B} and 87.61\% on \texttt{Qwen-7B} outperforming CoT by \textbf{+8.69\%} and \textbf{+28.74\%}, respectively. \textit{Unlike CoT, which confidently answers even in the absence of valid context, our agents are ``evidence-seeking", "input grounded", rather than ``answer-seeking," leading to more trustworthy and robust multimodal QA.}

\subsection{Choices in \methodName Architecture}

\paragraph{Dynamic vs. Static Agents}. Dynamic agentic frameworks like Tree-of-Thoughts (ToT)~\cite{10.5555/3666122.3666639} rely on explicit search typically via depth-first traversal to enumerate and rank multiple reasoning paths. In our setup, ToT is instantiated with 3 agents and a max depth of 3, generating an average of 12 thoughts per question. Despite this computational overhead, as shown in ~\autoref{tab:multimodal}, ToT achieves 57.12\% on \texttt{Qwen-7B}, while our static agentic method using only 3 sequential agents achieves \textbf{67.56\%}, a \textbf{+10.44\%} gain.

Beyond accuracy, ToT exhibits failure modes indicative of brittle search behavior: it frequently returns confidently incorrect answers (avg. confidence \textit{0.93}) and often declares multi-hop questions ``unanswerable," missing key compositional signals. In contrast, our static framework without iterative search or re-ranking demonstrates more grounded reasoning, better calibration, and robust handling of multi-modal, multi-hop queries. These findings challenge the assumption that dynamic search improves generalization, and highlight the efficacy of a lean, static agentic architecture in complex QA tasks.

\begin{figure}[!ht]
    \centering
    \includegraphics[width=\linewidth]{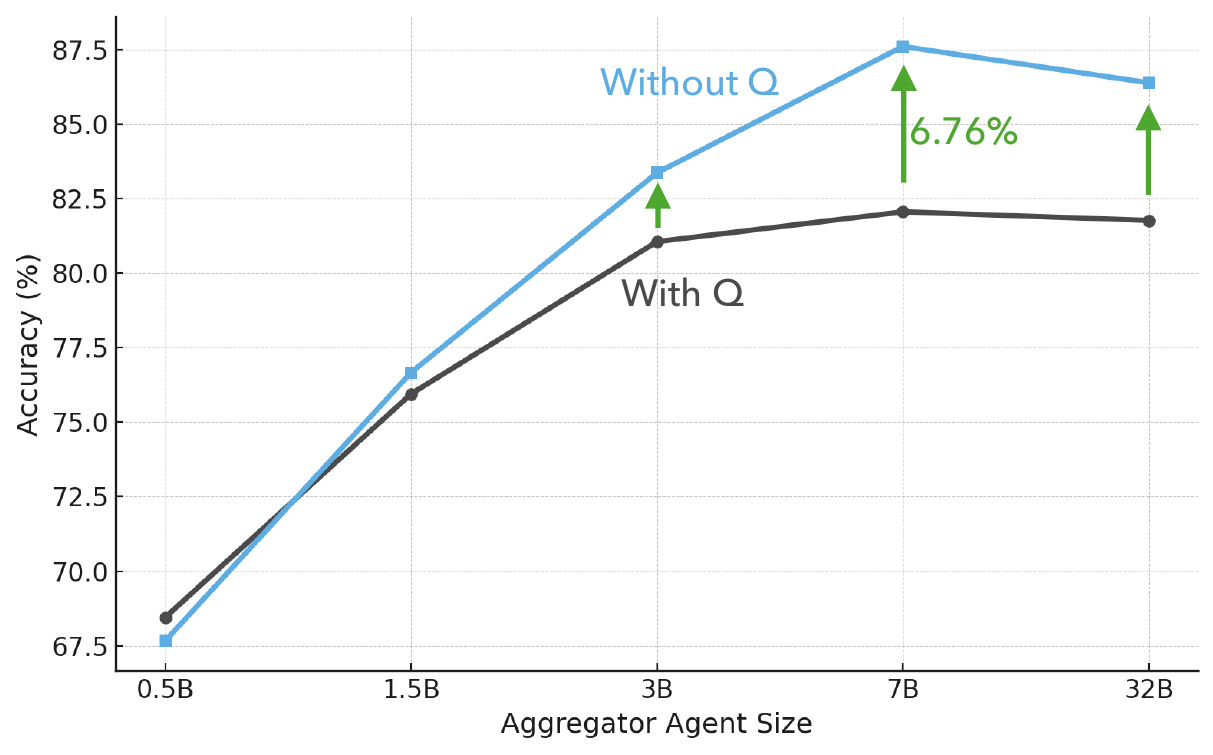}
    \caption{Aggregator Agent performance with and without question on \textsc{MultiModalQA}  Dataset.}
    \label{fig:agg_w_wo_q}
\end{figure}

\paragraph{Question Agnostic Aggregator.} As shown in ~\autoref{fig:agg_w_wo_q}, our experiments reveal withholding the original question from the Aggregator Agent consistently improves performance across model scales. For instance, at 7B parameters, accuracy increases from 82.06\% (with question) to \textbf{87.61\%} (without question). This improvement arises because removing the question reduces reliance on linguistic priors and compels the Aggregator to synthesize answers solely from cross modality-grounded evidence provided by expert agents. \textit{This, making the \methodName agent question agnostic make it's unbias, grounded, and more factual.}

Analogous to ensemble methods in NLP, where a meta-learner integrates outputs from base models without direct access to the input query, this separation mitigates bias and enhances answer faithfulness and factual consistency reducing hallucinations~\cite{puerto-etal-2023-metaqa}.

\section{Comparison with Related Work}

We position our work within four key areas of multimodal question answering (MMQA): benchmark design, unified models, prompt-based reasoning, and multi-agent systems. 

\paragraph{Benchmarks for Multimodal QA}
Datasets like \textsc{ManyModalQA}~\citep{Hannan_Jain_Bansal_2020} and \textsc{MultimodalQA}~\citep{talmor2021multimodalqa} challenge systems to reason across text, tables, and images, with a significant portion requiring cross-modal fusion. Others, such as \textsc{HybridQA}~\citep{chen-etal-2020-hybridqa}, \textsc{OTT-QA}~\citep{chen2021ottqa}, and \textsc{TAT-QA}~\citep{zhu2021tatqaquestionansweringbenchmark}, focus on structured-unstructured combinations. These benchmarks highlight core challenges in modality disambiguation and evidence integration issues \methodName{} explicitly tackles through specialized agents.

\paragraph{Unified Models}
Encoder-decoder architectures like \textsc{AutoRouting} and \textsc{ImplicitDecomp}~\citep{talmor2021multimodalqa} embed all modalities into a shared space. More advanced models \textsc{Binder}, \textsc{SKURG}, \textsc{PERQA}~\citep{yang2023progressiveevidencerefinementopendomain}, \textsc{Solar}~\citep{yu2023unifiedlanguagerepresentationquestion}, and \textsc{UniRaG}~\citep{sharifymoghaddam-etal-2025-unirag} incorporate retrieval and structural cues. While effective, these approaches often obscure modality roles and degrade with missing inputs. \methodName{} sidesteps these issues by activating only relevant agents per input.

\paragraph{Prompt-Based Reasoning}
Prompting strategies like Chain-of-Thought~\citep{wei2022chain} and its multimodal variants~\citep{zhang2023multimodal, zheng2023ddcot}, including Tree-of-Thoughts~\citep{10.5555/3666122.3666639}, offer zero-shot reasoning capabilities. However, they typically depend on a single LLM, making them prone to hallucinations and conflicts while \methodName{} distributes reasoning across modality-specific agents.

\paragraph{Multi-Agent Systems}
Agent-based approaches such as \textsc{ReConcile}~\citep{chen2023reconcile} leverage collective decision-making, often via voting. Though applied to math and planning~\citep{10.5555/3666122.3666639}, they remain underexplored in MMQA. \methodName adapts this paradigm by coordinating agents across modalities with structured synthesis, enabling verifiable and interpretable reasoning. 

Unlike prior approaches that rely on monolithic models or single-agent prompting, \methodName introduces a multi-agent, interpretable architecture. It assigns specialized roles to agents based on modality and separates reasoning into distinct stages: extraction, synthesis, and aggregation. This design enables \methodName to match or exceed the performance of state-of-the-art MMQA models while offering improved interpretability, robustness, and zero-shot generalization.

\section{Conclusion}

We present \textbf{\methodName}, a modular, prompt-driven multi-agent framework for multimodal QA that performs structured reasoning through modality-specific extraction, cross-modal synthesis, and evidence-grounded aggregation entirely without finetuning. \methodName achieves state-of-the-art zero-shot results on both \textsc{MultimodalQA} and \textsc{ManyModalQA}, outperforming prompting-based baselines and several finetuned models. On \textsc{MultimodalQA}, it achieves 76.37\% with \texttt{Qwen2.5-VL-7B} and 67.56\% with \texttt{Qwen2.5-VL-7B}, surpassing CapCoT and Tree-of-Thoughts by over 6\% and 10\%, respectively. On \textsc{ManyModalQA}, it reaches 89.90\% with \texttt{Qwen2.5-VL-7B} and 87.61\% with \texttt{Qwen2.5-VL-7B}, outperforming CoT by up to 28.74\%. In addition to strong performance, \methodName exhibits higher robustness and interpretability. It remains stable under irrelevant context and avoids hallucination in perturbed settings, while static agents outperform dynamic search-based methods like ToT with less complexity. These results underscore \methodName as a scalable, interpretable, and high-performing zero-shot solution for multimodal QA.

\section{Limitations} \methodName’s reliance on separate LLM/VLM experts for each modality simplifies zero-shot generalization but incurs substantial inference latency, memory usage, and monetary cost. Extending the framework to additional modalities (e.g., audio, video, sensor data) would require equally capable foundation models or complex preprocessing pipelines, limiting applicability in resource-constrained or real-time environments. The three-stage design (Aggregator) enhances transparency but makes the system brittle: mistakes in early extraction cannot be corrected downstream, and the Aggregator blind to raw inputs cannot recover missing or misinterpreted evidence. This fragility is reflected in our perturbation tests, where scrambled or incomplete context causes small accuracy drops. Incorporating iterative feedback or retrieval loops could improve robustness but would complicate the current prompt-driven simplicity.

\section{Ethics Statement}
The authors affirm that this work adheres to the highest ethical standards in research and publication. Ethical considerations have been meticulously addressed to ensure responsible conduct and the fair application of computational linguistics methodologies. Our findings are aligned with experimental data, and while some degree of stochasticity is inherent in black-box Large Language Models (LLMs), we mitigate this variability by maintaining fixed parameters such as temperature and top p for consistent generation across the models used. Furthermore, our use of LLMs, including \texttt{GPT-4o-mini}, \texttt{Gemini-1.5-flash-8B}, and \texttt{Qwen} models, complies with their respective usage policies. The research involved the analysis and correction of dataset mislabeling to ensure data integrity. We have used AI assistants (Grammarly and ChatGPT) to address the grammatical errors and rephrase the sentences. Finally, to the best of our knowledge, we believe that this work introduces no additional risk. To the best of our knowledge, this study introduces no additional ethical risks.

\section*{Acknowledgements}
We thank the Complex Data Analysis and Reasoning Lab at Arizona State University for computational support, and Vihang Pancholi for his help during the submission process.

\bibliography{custom}

\newpage
\appendix
\section{Additional Experimental Results}
\label{appendix:extended_results}

This section reports supplementary experiments extending our main evaluation. These include ablations analyzing model modularity, efficiency, and robustness.

\paragraph{Ablation on Synthesizer and Single Expert.}
Following reviewer feedback, we tested (a) removal of the \textit{Cross-Modal Synthesizer} and (b) replacement of modality-specialized experts with a \textit{single unified expert}. The results are presented in Table~\ref{tab:ablation_synth_expert}.

\begin{table}[h!]
\centering
\scriptsize
\setlength{\aboverulesep}{1.5pt}
\setlength{\belowrulesep}{1.5pt}
\setlength{\tabcolsep}{3pt}
\begin{tabular}{lcccc}
\toprule
\textbf{Model} & \textbf{MultiModalQA} & \textbf{Single} & \textbf{ManyModalQA} & \textbf{Single} \\
\textbf{} & \textbf{} & \textbf{Expert} & \textbf{} & \textbf{Expert} \\
\midrule
\texttt{Qwen2.5-VL-7B} & 76.37 & 72.17 & 89.90 & 85.70 \\
Gemini 8B & 65.84 & 61.64 & 87.91 & 83.71 \\
\texttt{Qwen2.5-VL-3B} & 67.56 & 53.36 & 87.61 & 63.41 \\
\bottomrule
\end{tabular}
\caption{Ablation analysis on the Synthesizer and single-expert variants. Removing the Synthesizer or unifying experts both reduce accuracy by 4-25 points, demonstrating the necessity of modular specialization.}
\label{tab:ablation_synth_expert}
\end{table}

Removing the cross-modal Synthesizer causes significant drops (up to 24.2 pp on multi-hop tasks with \texttt{Qwen-7B}). Using a single unified expert consistently under performs relative to modality-specialized agents, confirming that modular decomposition is crucial not over-engineering.

\paragraph{Latency and Cost Efficiency.}
Although MAMMQA involves multiple agents, each executes a single prompt, unlike CoT, which relies on multi-sample decoding. Consequently, MAMMQA remains both interpretable and cost-efficient. 

{
\vspace{-0.5em}
\small
\begin{itemize}[noitemsep]
    \item \texttt{Qwen} 2.5-VL 3B + MAMMQA: \textbf{86.37\%}  
    \item \texttt{Qwen} 2.5-VL 7B + MAMMQA: \textbf{87.61\%}  
\end{itemize}
\vspace{-0.5em}
}

Both outperform stronger baselines such as \texttt{Qwen-7B} + CoT (66.54\%) and Gemini 8B + CoT (84.81\%), as well as GPT-4o + CoT (86.20\%), demonstrating higher \textit{performance per parameter}. The agentic overhead is minimal compared to CoT sampling-based reasoning.

\paragraph{Semantic Sensitivity: Text Shuffle Test.}
We additionally perform a ``text shuffle" stress test, where words (not sentences) are randomly permuted to disrupt semantic coherence. CoT baselines show resilience (indicating reliance on superficial token associations), while MAMMQA suffers a larger drop ($\approx$91.2$\%$), highlighting stronger semantic sensitivity-an important trait for grounded multimodal reasoning.

These results empirically validate the design choices in MAMMQA: (i) modular specialization across experts is essential, (ii) the Synthesizer plays a key role in cross-modal composition, and (iii) the framework achieves both interpretability and cost efficiency without sacrificing accuracy.

\section{Qualitative Walkthrough Examples}
\label{appendix:qualitative_examples}

We further illustrate MAMMQA’s reasoning process through representative examples.



\begin{exampleblock}
\textit{\textbf{Example 1: ``Which animal in the image is a mammal?"}} \\
\textbf{Input:} An image showing a frog, a dolphin, and a bird, with the caption ``These animals live in different habitats."  \\
\textbf{Answer:} \textit{Dolphin}. 

\medskip
\textbf{Stage 1 -- Expert Agents:}  
Image agent identifies animal regions; text agent extracts relevant concepts (``different habitats"). 

\medskip    
\textbf{Stage 2 -- Synthesizer:}  
Fuses visual and textual evidence to reason over biological traits.

\medskip 
\textbf{Stage 3 -- Aggregator:}  
Produces the reasoning trace:  \texttt{[\textbf{Visual}: Dolphin → live birth] + [\textbf{Text}: warm-blooded] → Dolphin is mammal.}
\end{exampleblock}

\paragraph{Baseline (CoT) Comparison.}
CoT misclassifies ``bird" due to token frequency bias. MAMMQA’s modular design ensures factual grounding and interpretability.

\begin{exampleblock}
\textit{\textbf{Example 2: ``According to the chart and text, which city had the highest rainfall?"}}\\
\textbf{Input:} A bar chart with rainfall data and a paragraph on weather patterns. \\ 
\textbf{Answer:} \textit{Singapore}.
\end{exampleblock}

MAMMQA correctly identifies ``Rainfall (mm)" as the relevant dimension, grounds the chart bars, aligns textual cues (``tropical"), and aggregates them for final reasoning.

These examples demonstrate how MAMMQA provides transparent, step-wise multimodal reasoning, maintaining interpretability while achieving competitive quantitative performance.



\section{Prompts Details}
\label{sec:appendix_prompts}

\mytcbinputwide{prompts/agent_stage_1.tex}{Modality Expert Agent Prompt}{0}{bw:domain}{prompt:AS1}

\mytcbinputwide{prompts/agent_stage_2.tex}{Cross Modality Agent Prompt}{0}{bw:domain}{prompt:AS2}

\mytcbinputwide{prompts/agent_stage_3.tex}{Aggregator Prompt}{0}{bw:domain}{prompt:AS3}



\end{document}